\documentclass[runningheads]{llncs}
\usepackage[T1]{fontenc}
\usepackage{graphicx}
\usepackage{booktabs}  % For better table lines
\usepackage{xcolor}
\usepackage{cite}
\usepackage{array}     % For column alignment
\usepackage{marvosym}
\usepackage{adjustbox} % For adjusting box size
\usepackage{multirow}  % For multirow cells
\usepackage{float}     % For exact table placement
\usepackage{tabularx}  % For equal width columns
\usepackage{ragged2e}  % For left-aligning text in tabularx
\usepackage{hyperref}
\usepackage{CJKutf8}

\begin{document}
\begin{CJK*}{UTF8}{gbsn}
\title{Dialogue-Based Multi-Dimensional Relationship Extraction from Novels}
%
%\titlerunning{Abbreviated paper title}
% If the paper title is too long for the running head, you can set
% an abbreviated paper title here
\author{Yuchen Yan \and Hanjie Zhao \and Senbin Zhu\and Hongde Liu\and Zhihong Zhang\and Yuxiang Jia\textsuperscript{(\Letter)}}
\authorrunning{Y. Yan et al.}
% % First names are abbreviated in the running head.
% % If there are more than two authors, 'et al.' is used.
% %
\institute{School of Computer and Artificial Intelligence, Zhengzhou University, Zhengzhou, P.R.China\\
\email{\{yanyuchen,hjzhao\_zzu,nlpbin,lhd\_1013\}@gs.zzu.edu.cn},\\
\email{\{iezhzhang,ieyxjia\}@zzu.edu.cn}}
% \email{ieyxjia@zzu.edu.cn}
%
\maketitle              % typeset the header of the contribution
\begin{abstract}
Relation extraction is a crucial task in natural language processing, with broad applications in knowledge graph construction and literary analysis. However, the complex context and implicit expressions in novel texts pose significant challenges for automatic character relationship extraction. This study focuses on relation extraction in the novel domain and proposes a method based on Large Language Models (LLMs). By incorporating relationship dimension separation, dialogue data construction, and contextual learning strategies, the proposed method enhances extraction performance. Leveraging dialogue structure information, it improves the model's ability to understand implicit relationships and demonstrates strong adaptability in complex contexts. Additionally, we construct a high-quality Chinese novel relation extraction dataset to address the lack of labeled resources and support future research. Experimental results show that our method outperforms traditional baselines across multiple evaluation metrics and successfully facilitates the automated construction of character relationship networks in novels. 

\keywords{Novel Texts  \and Dialogue \and Character Relationship Extraction.}
\end{abstract}
\section{Introduction}
Novels, as one of the most popular forms of literature, are known for their rich narratives and vivid character interactions, which often reflect complex social relationships\cite{rp1}. However, effectively analyzing and understanding these character relationships in an automated manner remains a significant challenge in the field of Natural Language Processing\cite{rp2}. While recent advances in large-scale pre-trained language models have significantly improved text understanding, most relationship extraction methods still focus on structured domains like news or encyclopedias\cite{rp3}\cite{rp4}. In contrast, research on literary texts—especially novels characterized by intricate narrative structures and metaphorical expressions—remains relatively limited. Automatic extraction of character relationships from novels not only facilitates automated literary analysis but also supports applications in knowledge graph construction\cite{rp5}, intelligent recommendation, and digital humanities\cite{rp6}.

\begin{figure}
  \centering
  \includegraphics[width=0.8\linewidth]{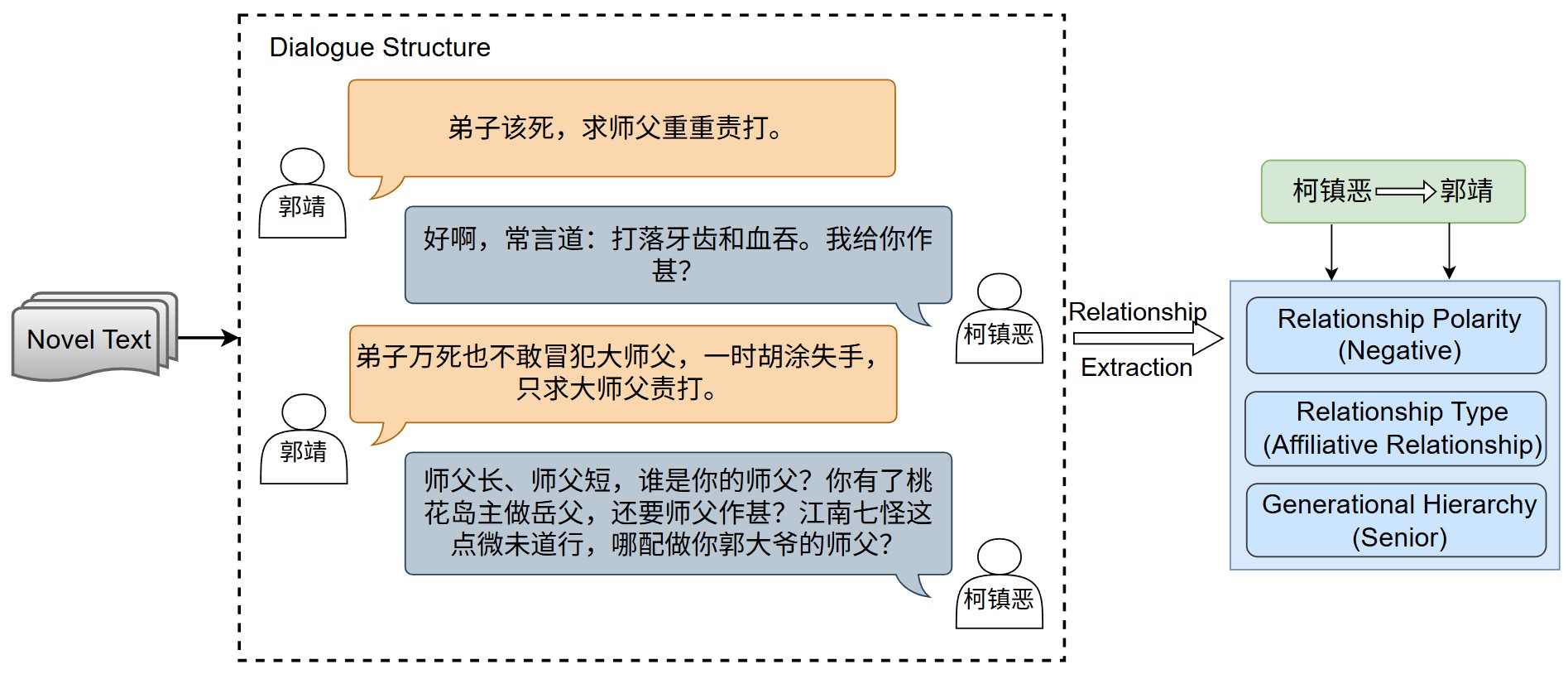}
  \caption{Character Relationship Extraction from Novels}
  \label{fig1}
\end{figure}

This study focuses on the automatic extraction of character relationships in novels, as shown in Figure \hyperref[fig1]{1}, and proposes a method based on LLMs. Unlike traditional relationship extraction methods, this work leverages dialogue structure and designs a multi-dimensional relationship extraction framework to address the implicit and complex character relationships in novels. Specifically, the proposed approach introduces a relationship dimension separation strategy, explores the construction of dialogue-based data, and employs a retrieval-based in-context learning strategy combined with parameter-efficient fine-tuning techniques to enhance task performance while maintaining computational efficiency. To support the research, a high-quality Chinese novel character relationship extraction dataset was constructed. Experimental results demonstrate that the proposed method significantly outperforms traditional baselines across multiple evaluation metrics and is successfully applied to the automatic construction of novel-based social relationship networks, further validating its applicability to literary analysis tasks.

The main contributions of this paper are summarized as follows:
\begin{enumerate}
\item[$\bullet$]A multi-dimensional character relationship extraction framework is proposed, which recognizes relationship types based on dialogue structure and incorporates a LLMs-based task optimization strategy,  improve model adaptability to complex novel texts.
\item[$\bullet$]A high-quality Chinese novel character relationship extraction dataset is constructed, providing reliable experimental data for future research and promoting the development of this field. The corpus will be made publicly available at https://github.com/LimboChen/NCRE-dataset.
\item[$\bullet$]The effectiveness of the proposed method is validated both experimentally and in application. The method not only outperforms traditional baselines in multiple evaluation metrics but also proves effective in the automated construction of social relationship networks in novels, offering new perspectives for digital humanities research.
\end{enumerate}	

\section{Related Work} 
Relation extraction is a fundamental task in the field of Natural Language Processing (NLP), aiming to identify semantic relationships between entities from unstructured text. Applying relation extraction techniques to novel analysis enables the automatic identification of various interpersonal relationships, which not only facilitates large-scale text analysis but also provides valuable data for the construction and interpretation of character social networks.

In the domain of literary works, existing research on character relationship extraction has mainly focused on screenplays of film and television productions. These works leverage their clearly defined dialogue structures and multi-turn interactions to identify relationships between speakers, typically modeled through multi-turn {speaker: utterance} patterns. Several high-quality datasets have been constructed in this field, supporting a range of studies. An overview of representative corpora is presented in Table 1.

\begin{table}
\centering
% \captionsetup{justification=centering}
\caption{Dialogue Relation Extraction Corpus}
\label{tab1}
\begin{tabular}{>{\centering\arraybackslash}p{2cm} >{\centering\arraybackslash}p{2.5cm} >{\centering\arraybackslash}p{3.5cm} >{\centering\arraybackslash}p{3.5cm}}
    \toprule
    Corpus & Language &  \#Relationship Types & \#Relationship Labels \\
    \toprule
    FiRe\cite{rp7} & English & 12 & 783 \\
    DialogRE\cite{rp8} & English\&Chinese &  36 & 9,773 \\
    CRECIL\cite{rp9} & Chinese &  30 & 45,874 \\
    \bottomrule
\end{tabular}
\end{table}

Yu et al.\cite{rp8} introduced the DialogRE dataset, derived from the script of Friends, and formally proposed the task of relation extraction from dialogues. DialogRE defines 36 relation types and has become a widely used benchmark\cite{rp10}. \cite{rp11} In recent work, Li et al.\cite{rp12} explored the potential of using LLMs, such as ChatGPT, for relation extraction on the DialogRE dataset. They designed multiple prompt templates to guide ChatGPT in completing the task and applied both prompt tuning and fine-tuning methods to open-source models. Their results demonstrate that LLMs exhibit notable advantages in handling complex contextual information within dialogue relation extraction tasks.

Tigunova et al.\cite{rp7} proposed the PRIDE model for character relationship prediction in dialogues. They first constructed two datasets based on movie and TV scripts, and then integrated reinforcement learning with deep learning techniques to model contextual information in dialogues. Their method successfully captured latent relationship patterns between characters and demonstrated the potential of deep learning in this task.

In the Chinese dialogue domain, Jiang et al.\cite{rp9} developed the CRECIL dataset based on Chinese sitcom scripts. CRECIL focuses on automatically extracting character relationships in multi-party dialogues and includes annotations of global character relations and coreference links to generate relationship triples. Building on this dataset, Li et al.\cite{rp13} proposed a contrastive prompt tuning method (CoPrompt), which learns better relation embeddings by constructing positive and negative prompt pairs. Their approach achieved strong performance on both the DialogRE and CRECIL datasets, particularly setting new benchmarks on Chinese relation extraction tasks.

\section{Corpus Construction}
\subsection{Annotation Process and Schema}
\begin{figure}
  \centering
  \includegraphics[width=1\linewidth]{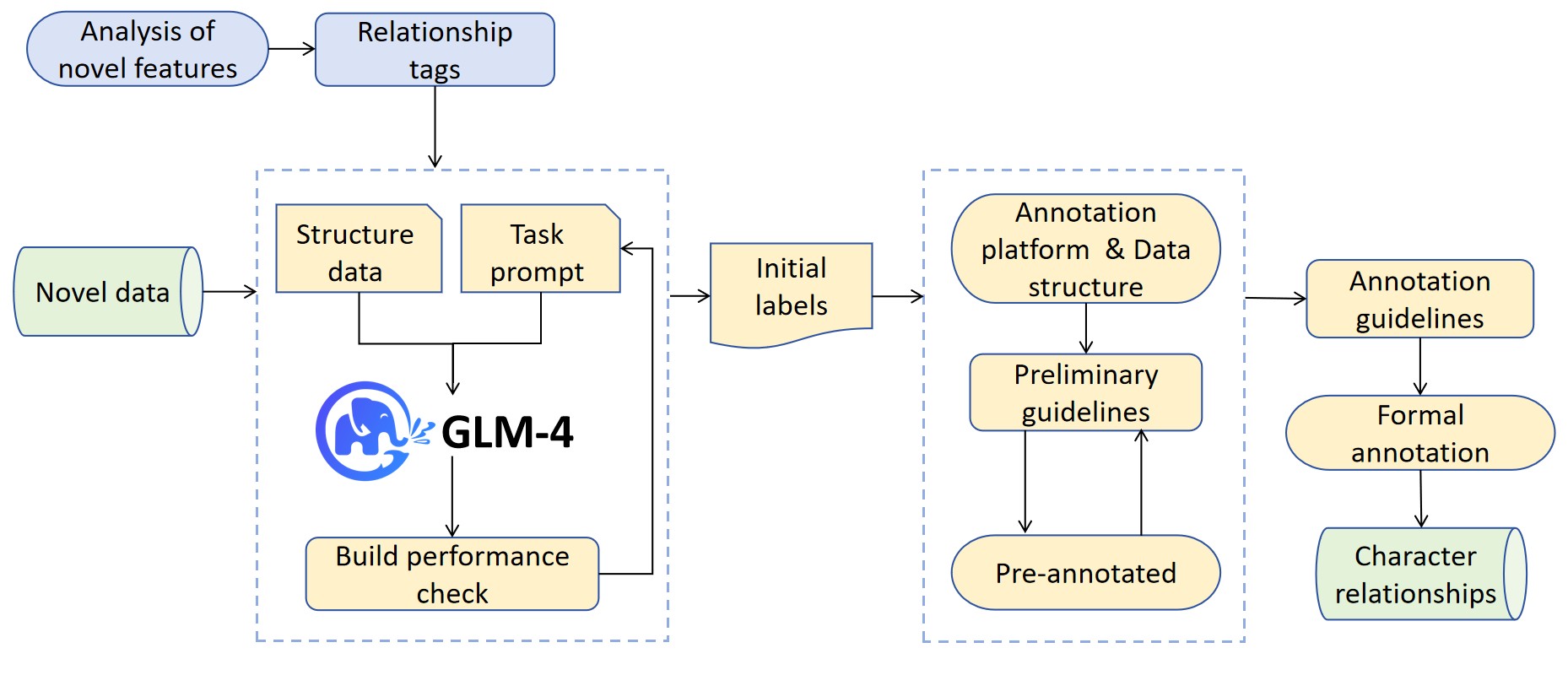}
  \caption{Annotation Process of Character Relationship Data}
  \label{fig2}
\end{figure}

Prior to conducting research on character relationship extraction in novels, we constructed a Chinese dataset based on Jin Yong’s classic martial arts novel \textit{The Legend of the Condor Heroes}. The annotation task focused on identifying the relationships between specified character pairs based on the context of quoted speech. The overall annotation process is illustrated in Figure \hyperref[fig2]{2}.

The core objective of the annotation process was to determine the relationship between two specified characters through contextual understanding of the surrounding paragraph. For each instance, annotators selected the most appropriate relationship label across three parallel dimensions. These dimensions provide a multi-layered framework for relationship analysis, ensuring both comprehensiveness and granularity in the annotation results.

In contrast to conventional relation extraction tasks that employ fine-grained relational labels, our label schema was tailored to the unique characteristics of Jin Yong’s martial arts fiction. We defined the following three parallel dimensions to construct a more holistic view of character relationships:
\begin{enumerate}
\item[$\bullet$]Relationship Polarity: Reflects the emotional stance between characters in the given context, with labels including positive (friendly), neutral, and negative (hostile).
\item[$\bullet$]Relationship Type: Describes the social nature of the relationship, categorized as kinship (e.g., family, clan, sworn siblings), affiliative (e.g., workplace, mentorship, sect membership), and other relationships.
\item[$\bullet$]Generational Hierarchy: Represents the relative age or generational status between characters, labeled as senior, peer, or junior.
\end{enumerate}	

In the early stages of annotation, we incorporated the ChatGLM-4 model to perform preliminary labeling and generate candidate tags based on its natural language understanding capabilities. This assisted in improving annotation efficiency and reducing the initial manual workload. Following this, the research team conducted an in-depth analysis of the data and candidate annotations, established systematic annotation guidelines, and organized a trial annotation phase with all annotators.

During the formal annotation stage, annotators followed the refined annotation guidelines and carried out two rounds of iterative labeling and review. This iterative verification strategy helped ensure both the accuracy and consistency of the annotations. It is worth noting that selecting the appropriate relationship label often required combining prior narrative knowledge with the local context. For example, polarity labels typically relied on the current scene, whereas generational hierarchy labels were inferred from the broader background of the novel.

\subsection{Corpus Analysis}
The Novel Character Relationship Extraction (NCRE) dataset includes 100 characters from \textit{The Legend of the Condor Heroes}, comprising 1,109 dialogue units and 3,591 relationship instances, each annotated across three dimensions, totaling 10,773 relation labels. Statistical analysis shows that in the relation polarity dimension, positive (friendly) relations dominate (58.12\%), while negative relations also hold a significant portion (31.83\%), reflecting the coexistence of alliance and conflict in the narrative. In the age/generation dimension, peer-level relationships are most frequent (47.28\%), highlighting the equal-status interactions typical of a martial arts setting.
\begin{figure}
  \centering
  \includegraphics[width=1\linewidth]{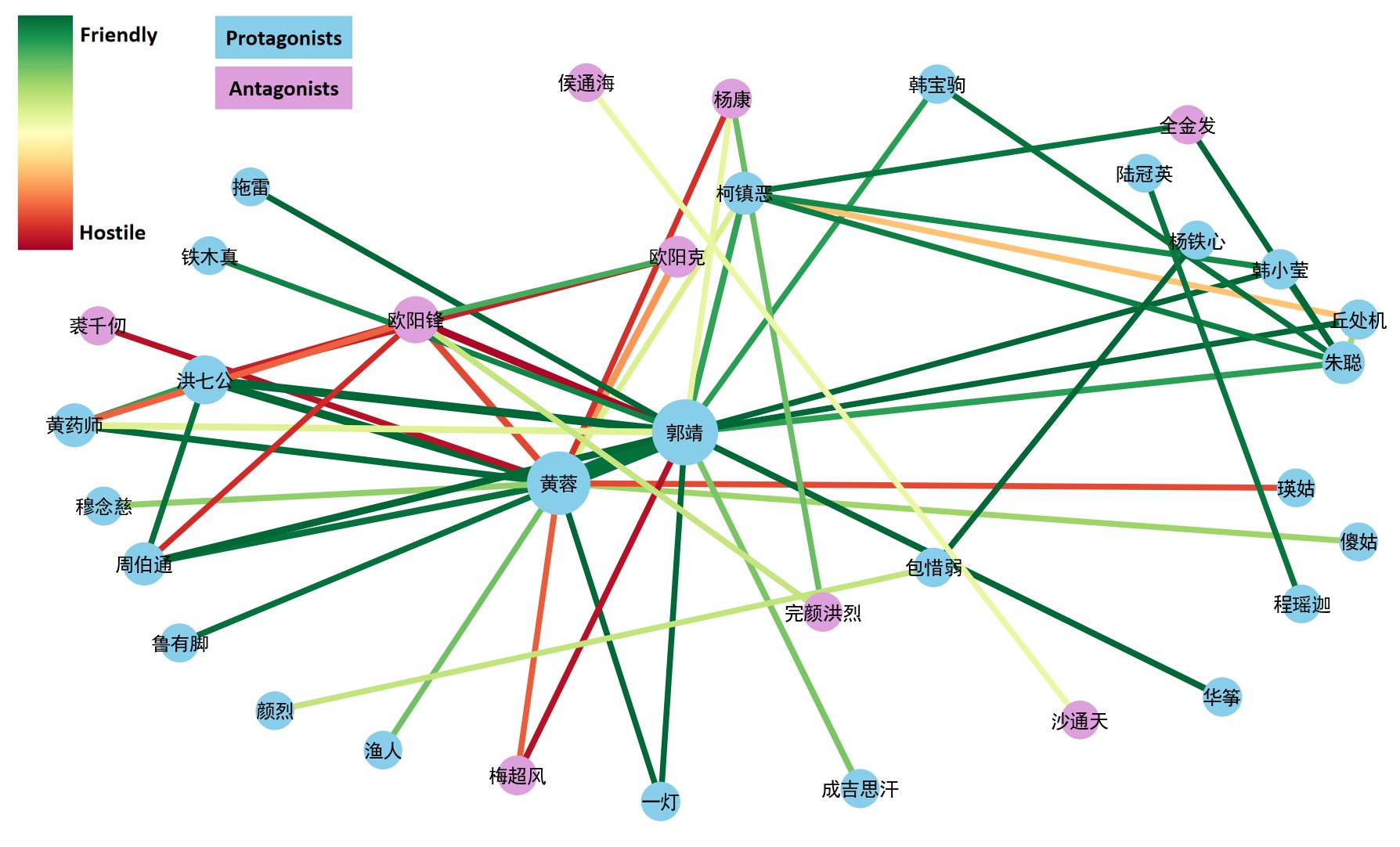}
  \caption{Friendship-Hostility Relationship Network of Characters in Novel}
  \label{fig3}
\end{figure}

In the field of automated literary analysis, constructing a character relationship network facilitates the intuitive representation of social connections and plot interactions among characters. Based on the NCRE corpus, this study constructs the character relationship network for \textit{The Legend of the Condor Heroes}. In the network, nodes represent major characters, with node size determined by the smoothed frequency of direct quotations, reflecting the character’s prominence in the narrative. The thickness of edges between nodes indicates the frequency of dialogue interactions, revealing the intensity of interpersonal connections. Furthermore, the network incorporates a relation polarity dimension to build an undirected graph based on emotional polarity. The color of edges ranges from red to green, indicating the strength of hostile to friendly relationships, while node colors distinguish between protagonists and antagonists, enhancing both readability and the interpretability of character roles.

The character relationship network reveals the overall structure and features of interpersonal relationships in the novel, as shown in Figure \hyperref[fig3]{3}. Analysis shows that Guo Jing and Huang Rong have the most frequent interactions, consistently maintaining a highly friendly relationship, aligning with their emotional arc in the story. The edges between the protagonists and various antagonists are marked in deep red, reflecting intense opposition. Some character relationships exhibit transitional or asymmetric patterns—for instance, the relationship between Guo Jing and Yang Kang reflects a weighted combination of friendship and hostility, while the interaction between Huang Rong and Ouyang Ke indicates a blend of one-sided friendliness (from Ouyang Ke) and hostility (from Huang Rong). The structure and visual encoding of the network not only assist in identifying key characters and major conflicts but also highlight the diversity and dynamic evolution of character relationships, offering valuable data support for in-depth literary analysis.

\section{Multi-dimensional Character Relationship Extraction} 
\subsection{Methodology}
\begin{figure}
  \centering
  \includegraphics[width=1\linewidth]{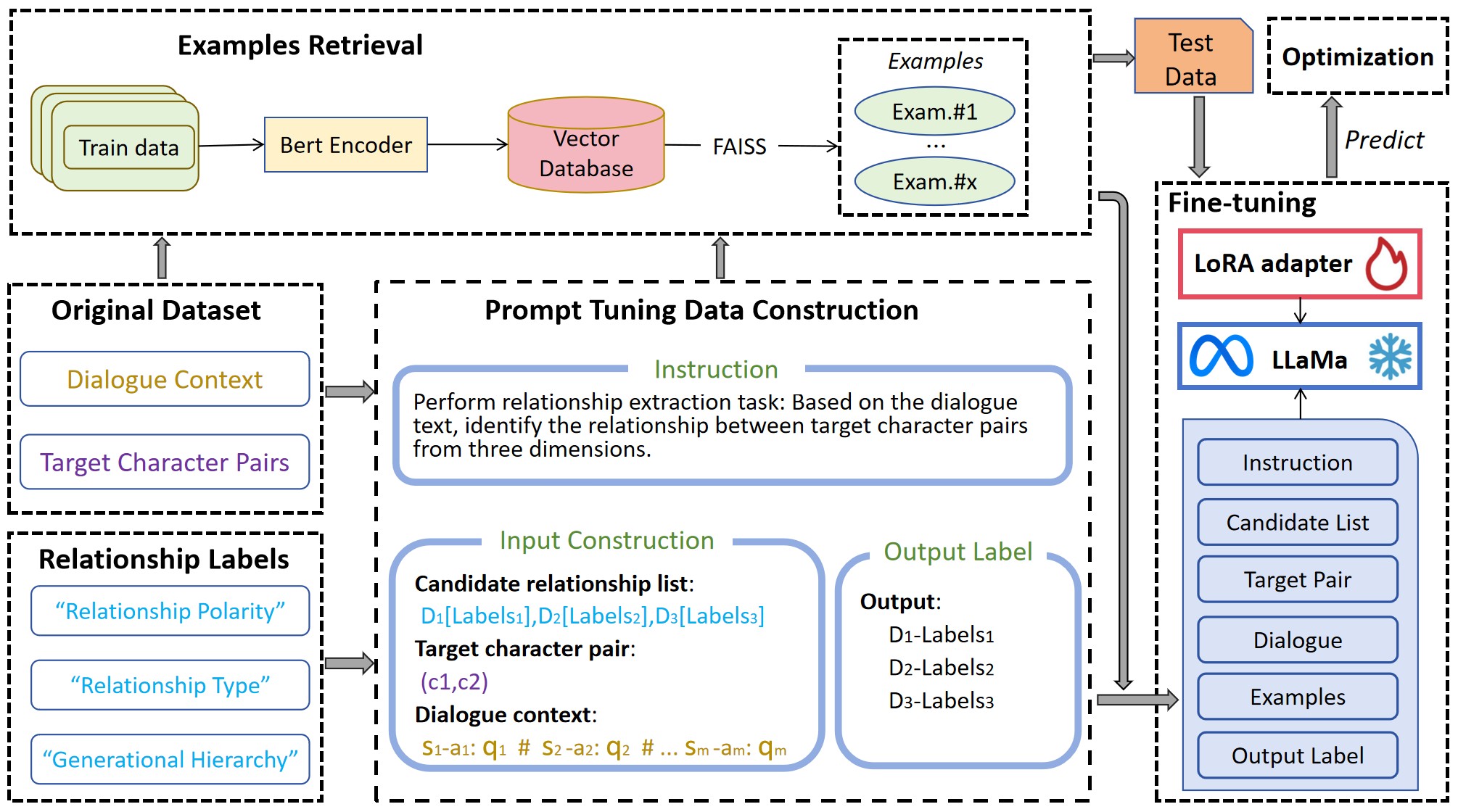}
  \caption{Character Relationship Extraction with Dialogue Information}
  \label{fig4}
\end{figure}

We propose CREDI (Character Relationship Extraction with Dialogue Information), a method for character relationship extraction based on dialogue structure information.The overall structure of the proposed model is shown in Figure \hyperref[fig4]{4}. Built upon the open-source LLaMa 3.1 model\cite{rp14}, CREDI integrates strategies such as dialogue data construction, relationship dimension separation, prompt optimization, example retrieval, and efficient parameter tuning to enhance extraction accuracy and robustness.
\subsubsection{Relationship Category Division} To address issues of fine-grained and overlapping relationship labels in traditional methods, CREDI reorganizes relationships into three parallel dimensions with coarse-grained labels, providing a comprehensive perspective for better relationship understanding.
\subsubsection{Dialogue Data Construction} An "Expanded Dialogue" strategy is proposed, extracting speaker and listener information from novel texts and reconstructing dialogues in the format "A said to B," thereby clarifying interaction structures and improving the model’s perception of character interactions.
\subsubsection{Prompt Optimization} Structured prompts are designed to encode the target characters, candidate labels, and dialogue information, guiding the model to focus on key information and improving the consistency and accuracy of relationship extraction.
\subsubsection{Retrieval-Based In-Context-Learning} Using multilingual BERT to vectorize the training data and FAISS\cite{rp15} for Top-K similarity retrieval, similar examples are incorporated as contextual prompts, significantly enhancing the model's semantic understanding and generation of character relationships.
\subsubsection{Parameter-Efficient Fine-Tuning} LoRA\cite{rp16} technology is applied to perform lightweight fine-tuning on LLaMa 3.1, primarily targeting the Query and Value matrices. This approach preserves pre-trained knowledge while reducing computational costs and improving task adaptability.

% {\bfseries Relationship Category Division:} To address issues of fine-grained and overlapping relationship labels in traditional methods, CREDI reorganizes relationships into three parallel dimensions with coarse-grained labels, providing a comprehensive perspective for better relationship understanding.

% {\bfseries Dialogue Data Construction:} An "Expanded Dialogue" strategy is proposed, extracting speaker and listener information from novel texts and reconstructing dialogues in the format "A said to B," thereby clarifying interaction structures and improving the model’s perception of character interactions.

% {\bfseries Prompt Optimization:} Structured prompts are designed to encode the target characters, candidate labels, and dialogue information, guiding the model to focus on key information and improving the consistency and accuracy of relationship extraction.

% {\bfseries Retrieval Based In-Context-Learning:} Using multilingual BERT to vectorize the training data and FAISS for Top-K similarity retrieval, similar examples are incorporated as contextual prompts, significantly enhancing the model's semantic understanding and generation of character relationships.

% {\bfseries Parameter-Efficient Fine-Tuning:} LoRA technology is applied to perform lightweight fine-tuning on LLaMa 3.1, primarily targeting the Query and Value matrices. This approach preserves pre-trained knowledge while reducing computational costs and improving task adaptability.
\subsection{Datasets and Baselines}
The primary experiments are conducted on the Chinese novel dataset NCRE, with further generalization experiments performed on three public datasets: FiRe, DialogRE, and CRECIL. Each dataset is split into training, validation, and test sets at a ratio of 8:1:1 (following the original splits for CRECIL and DialogRE). To prevent knowledge leakage, character names in the DIR2E training set are randomly replaced with unique codes. For CRECIL and DialogRE, due to highly imbalanced label distributions, label threshold filtering and random sampling strategies are applied to balance the data and reduce dataset size, which has been proven effective through multiple experiments.

Three types of baseline methods are employed, representing different technical approaches to relation extraction:

\begin{enumerate}
\item[$\bullet$]{\bfseries Traditional Classification Method (BERT\_Classification):} Formulates relation extraction as a classification task using the BERT model, where the hidden vector of the [CLS] token serves as the semantic representation of the input, followed by a fully connected layer for classification.
\item[$\bullet$]{\bfseries Fine-tuned Pre-trained Models (T5):} Redefines relation extraction as a generative reading comprehension task based on fine-tuned pre-trained models (e.g., T5). By designing task-specific prompts, the model generates target relations, leveraging the generative framework and contextual understanding to enhance flexibility and performance.
\item[$\bullet$]{\bfseries Large Language Model (GPT-3.5):} Uses the GPT-3.5 Turbo API for batch prediction, employing prompt learning strategies to guide the model in understanding task requirements and generating outputs. Both zero-shot and few-shot settings are explored, with the few-shot setup providing three examples to improve task familiarity and prediction accuracy.
\end{enumerate}	

\subsection{Experiments}
Weighted-F1 was used as the evaluation metric to address class imbalance. As shown in Table 2, the fine-tuned large model outperformed baseline methods across all three relation dimensions on the NCRE dataset, achieving an F1 score of 0.79 in the relation polarity dimension. This demonstrates the model’s strength in complex semantic inference on novel texts. Furthermore, fine-tuning and strategy optimization significantly improved performance over direct use of GPT-3.5. Few-shot learning showed some improvement but exhibited instability due to sample selection bias.

\begin{table}
\centering
% \captionsetup{justification=centering}
\caption{Experimental Results on novel Character Relationship Extraction}
\label{tab2}
\begin{tabular}{>{\centering\arraybackslash}p{1.6cm} >{\centering\arraybackslash}p{1.6cm} >{\centering\arraybackslash}p{2.5cm} >{\centering\arraybackslash}p{2.5cm} >{\centering\arraybackslash}p{2.5cm}}
    \toprule
    \multicolumn{2}{c}{Model}
    & Relationship Polarity & Relationship Type & Generational Hierarchy \\
    \midrule
    \multicolumn{2}{c}{Fine-tuned PTMs}
    & 0.64 & 0.43 & 0.62 \\
    \multicolumn{2}{c}{Bert\_classification}
    & 0.65 & 0.44 & 0.42 \\
    \multirow{2}{*}{GPT-3.5}
    & zero-shot & 0.52 & 0.26 & 0.36 \\
    & few-shot & 0.69 & 0.35 & 0.31 \\
    \multicolumn{2}{c}{\textbf{CREDI}(Ours)}
    & \textbf{0.79} & \textbf{0.76} & \textbf{0.74} \\
    \bottomrule
\end{tabular}
\end{table}

To validate the effectiveness of the optimization strategies, ablation studies were conducted on relation type extraction and dialogue data construction, with results shown in Table 3. Multi-dimensional relation extraction outperformed single-dimension settings by providing a more comprehensive understanding of text information. Additionally, expanding dialogue data by introducing addressees information significantly enhanced the model’s ability to understand and extract dialogue semantics, confirming the importance of complete dialogue structures.
\begin{table}
\centering
% \captionsetup{justification=centering}
\caption{Ablation Study Results}
\label{tab3}
\begin{tabular}{>{\centering\arraybackslash}p{4cm} >{\centering\arraybackslash}p{2.5cm} >{\centering\arraybackslash}p{2.5cm} >{\centering\arraybackslash}p{2.5cm}}
    \toprule
    Model & Relationship Polarity & Relationship Type & Generational Hierarchy  \\
    \toprule
    \textbf{CREDI} & \textbf{0.79} & \textbf{0.76} & \textbf{0.74} \\
    Single-dimension extraction & 0.77 &  0.66 & 0.68 \\
    Basic dialogue information & 0.73 & 0.70 & 0.67 \\
    \bottomrule
\end{tabular}
\end{table}

Experimental results on the CRECIL, FiRe, and DialogRE public datasets are shown in Table 4. The proposed CREDI method consistently outperformed both baseline methods and existing SOTA approaches, further demonstrating the effectiveness of fine-tuned large models. In contrast, traditional methods performed poorly on datasets with fine-grained relation labels, highlighting their limitations in complex relation modeling.
\begin{table}
\centering
% \captionsetup{justification=centering}
\caption{Comparison of Experimental Results on Public Datasets}
\label{tab4}
\begin{tabular}{>{\centering\arraybackslash}p{1.6cm} >{\centering\arraybackslash}p{1.6cm} >{\centering\arraybackslash}p{1.5cm} >{\centering\arraybackslash}p{1.5cm} >{\centering\arraybackslash}p{1.5cm}}
    \toprule
    \multicolumn{2}{c}{Model}
    & CRECIL & FiRe & DialogRE \\
    \midrule
    \multicolumn{2}{c}{Fine-tuned PTMs}
    & - & 0.28 & - \\
    \multicolumn{2}{c}{Bert\_classification}
    & 0.04 & 0.36 & 0.06 \\
    \multirow{2}{*}{GPT-3.5}
    & zero-shot & 0.07 & 0.49 & 0.16 \\
    & few-shot & 0.09 & 0.61 & 0.20 \\
    \multicolumn{2}{c}{SOTA}
    & 0.62 & 0.38 & 0.79 \\
    \multicolumn{2}{c}{\textbf{CREDI}(Ours)}
    & \textbf{0.63} & \textbf{0.71} & \textbf{0.80} \\
    \bottomrule
\end{tabular}
\end{table}
\subsection{Case Study}
This section presents a typical case to compare the performance of different models in complex contexts for character relationship extraction. For clarity, the case truncates lengthy context, highlights target characters in green and blue, and marks incorrect predictions in red for intuitive presentation.
\begin{table}
\centering
% \captionsetup{justification=centering}
\caption{Prediction Cases}
\label{tab5}
\begin{tabular}{>{\RaggedRight\arraybackslash}p{3cm} >{\RaggedRight\arraybackslash}p{3cm}>{\RaggedRight\arraybackslash}p{3cm}>{\RaggedRight\arraybackslash}p{3cm}}
   \toprule
    \multicolumn{4}{p{12cm}}{目标人物：（朱聪 —> 郭靖）
\textcolor{green}{\textbf{朱聪}}这一指虽是未用全力，但竟被他内劲化开，不禁更是惊讶，同时怒气大盛，喝道：“这还不是内功吗？”\textcolor{blue}{\textbf{郭靖}}心念一动：“难道那道长教我的竟是内功？”说道：“这两年来，有一个人每天晚上来教弟子呼吸、打坐、睡觉。弟子一直依着做，觉得倒也有趣好玩。不过他真的没传我半点武艺。他叫我千万别跟谁说。弟子心想这也不是坏事，又没荒废了学武，因此没禀告恩师。”说着跪下来磕了个头，道：“弟子知错啦，以后不敢再去跟他玩了。”六怪面面相觑，听他语气恳挚，似乎不是假话。韩小莹道：“你不知道这是内功吗？”郭靖道：“弟子真的不知道甚么叫做内功......” } \\
    \toprule
    Ground Truth & Negative & Affiliative & Senior   \\
    Fine-tuned T5 & Negative & \textcolor{red}{Kinship} & \textcolor{red}{Junior}   \\
    Bert\_classification & \textcolor{red}{Neutral} & \textcolor{red}{Other relationship} & \textcolor{red}{Junior}  \\
    GPT-3.5 & \textcolor{red}{Neutral} & Affiliative & Senior  \\
    CREDI & Negative & Affiliative & Senior  \\
    \bottomrule
\end{tabular}
\end{table}

The case is from \textit{The Legend of the Condor Heroes}, depicting a master-apprentice conflict where Master Zhu Cong scolds his disciple Guo Jing out of surprise and anger. The text reflects overt hostility, implicit professional ties, and hierarchical status, covering multiple relational dimensions.
In prediction results, the Fine-tuned T5 accurately identified the hostile emotion, GPT-3.5 performed well in recognizing the master-apprentice relationship and hierarchical positioning, while the CREDI model successfully captured all three key relational aspects, demonstrating a more comprehensive understanding and prediction capability.

\subsection{Applications of Character Relation Extraction}
Based on character relationship extraction, this section automates the construction of character social networks in novels to validate the effectiveness and application feasibility of the proposed method. Besides The \textit{The Legend of the Condor Heroes}, three additional novels—\textit{Heaven Sword and Dragon Saber}, \textit{Dao Sheng}, and \textit{The Sentimental Swordsman, the Ruthless Sword}—are selected to ensure diversity in author or genre.

The construction follows a pipeline approach, consisting of data preprocessing, dialogue chain segmentation, model prediction, and automatic network building. Separate social networks are generated based on different relationship dimensions to assist in character analysis.

\begin{figure}
  \centering
  \includegraphics[width=1\linewidth]{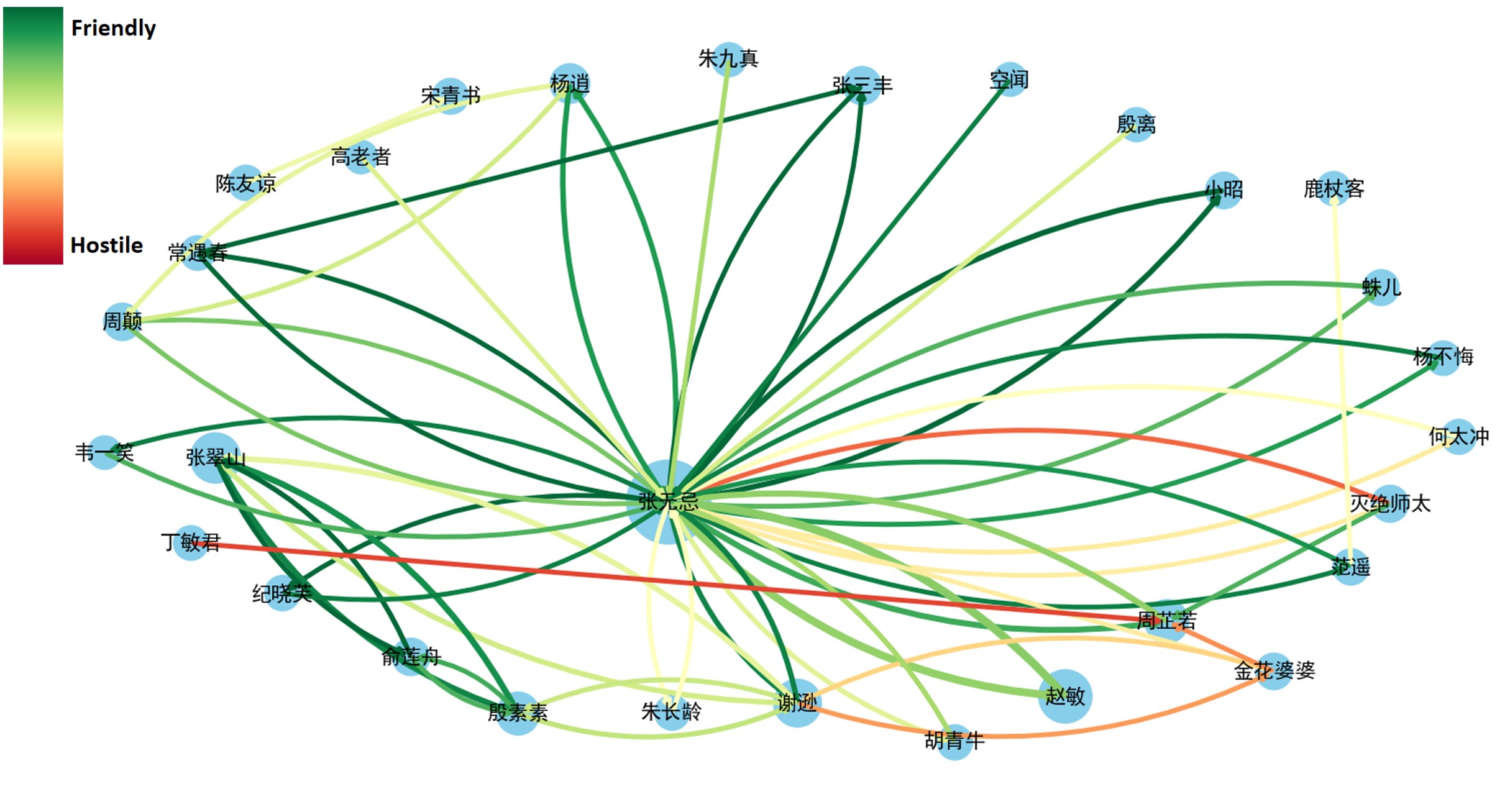}
  \caption{Character Relationship Network in \textit{Heaven Sword and Dragon Saber}}
  \label{fig5}
\end{figure}

Taking Heaven Sword and Dragon Saber as an example, as shown in Figure \hyperref[fig5]{5}, the Friendship-Hostility Relationship Network shows that protagonist Zhang Wuji maintains generally friendly ties with most characters, reflecting his tolerant and kind nature. His relationship with Zhao Min is complex, shown as light green, indicating occasional conflicts. Intense hostility among secondary characters highlights the chaotic factional struggles in the story. A typical case is Zhang Wuji’s restrained hostility toward enemy Miejue Shitai, showcasing his respectful and magnanimous character even towards adversaries.

\section{Conclusion} 
This paper aims to achieve automated extraction of multi-dimensional character relationships in novels. To extend the scope of relation extraction to literary texts, we construct a Chinese corpus for character relationship extraction, providing a reliable data foundation for information extraction tasks in the literary domain. We propose a character relationship extraction approach with dialogue structure, integrating multiple optimization strategies to enhance the performance of large language models on the task. Experimental results demonstrate the effectiveness of the proposed method. Furthermore, we explore its practical application by constructing automated character relationship networks in novels, verifying the feasibility of our approach in real-world literary analysis scenarios.

In future work, we plan to extend our research to a broader range of literary genres and formats. We also aim to develop more in-depth models for character relationship analysis, focusing on capturing the dynamic evolution of relationships throughout the narrative. 

\begin{credits}
\subsubsection{\discintname} The authors have no competing interests to declare that are relevant to the content of this article.
\end{credits}

\bibliography{myref.bib} %.bib文件名字
	
\bibliographystyle{splncs04} %.bst模板

\end{CJK*}
\end{document}